\documentclass{article}

\usepackage{booktabs}
\usepackage{pifont}
\usepackage{xcolor}
\usepackage{colortbl}
\usepackage{array}
\usepackage{subcaption}
\usepackage{graphicx} 
\usepackage{amsmath}
\usepackage{longtable}
\usepackage{enumitem}
\usepackage{algorithmicx}
\usepackage{algcompatible}
\usepackage{algpseudocode}

\usepackage[preprint]{neurips_2026}

\usepackage[utf8]{inputenc} 
\usepackage[T1]{fontenc}    
\usepackage{hyperref}       
\usepackage{url}            
\usepackage{booktabs}       
\usepackage{amsfonts}       
\usepackage{nicefrac}       
\usepackage{microtype}      
\usepackage{xcolor}         
\usepackage{wrapfig}
\usepackage{marvosym} 
\usepackage{algorithm}

\newcommand{\emailmark}{\textsuperscript{\Letter}}

\usepackage{xspace}
\makeatletter
\DeclareRobustCommand\onedot{\futurelet\@let@token\@onedot}
\def\@onedot{\ifx\@let@token.\else.\null\fi\xspace}
\def\eg{\emph{e.g}\onedot}

\makeatother

\title{ManipArena: A Controlled Benchmark for 
Diagnosing Generalization in Real-Robot Manipulation}

\author{
    \textbf{Anonymous Authors}
}

\author{
\textbf{Yu Sun}$^{1,2,*}$ \quad
\textbf{Meng Cao}$^{3,*}$ \quad
\textbf{Yang Ping}$^{2,*}$ \quad
\textbf{Kaidong Zhang}$^{1,2}$ \quad
\textbf{Qingxuan Chen}$^{2,5}$ \\
\textbf{Rongtao Xu}$^{3}$ \quad
\textbf{Liangwang Ruan}$^{2}$ \quad
\textbf{Xuecheng Chen}$^{2,4}$ \quad
\textbf{Dongxiu Liu}$^{2,4}$ \quad
\textbf{Yunxiao Yan}$^{2}$ \\
\textbf{Zunnan Xu}$^{4}$ \quad
\textbf{Runze Xu}$^{4}$ \quad
\textbf{Charles Yang}$^{2}$ \quad
\textbf{Peilun Zhang}$^{1}$ \quad
\textbf{Xiaofan Li}$^{2}$ \\
\textbf{Ruyi Gan}$^{2,\dagger}$ \quad
\textbf{Liang Ma}$^{3}$ \quad
\textbf{Yuehao Yin}$^{2}$ \quad
\textbf{Jincheng Yu}$^{4}$ \quad
\textbf{Lufang Chen}$^{2}$ \\
\textbf{Yuxin Liang}$^{2}$ \quad
\textbf{Peng Zhai}$^{2}$ \quad
\textbf{Hao Wang}$^{2}$ \quad
\textbf{Ivan Laptev}$^{3}$ \quad
\textbf{Ian Reid}$^{3}$ \\
\textbf{Qian Wang}$^{2}$\emailmark \quad
\textbf{Xiaodan Liang}$^{1,3}$\emailmark
\\[0.4em]
$^{1}$Sun Yat-sen University \quad
$^{2}$X Square Robot \quad
$^{3}$MBZUAI \quad
$^{4}$Tsinghua University \quad
$^{5}$University of Zurich
\\[0.2em]
$^{*}$\,Equal Contribution \quad $^{\dagger}$\,Project Leader \quad \emailmark\ Corresponding Authors
}

\begin{document}

\maketitle


\begin{figure}[th!]
    \centering
    \includegraphics[width=\linewidth]{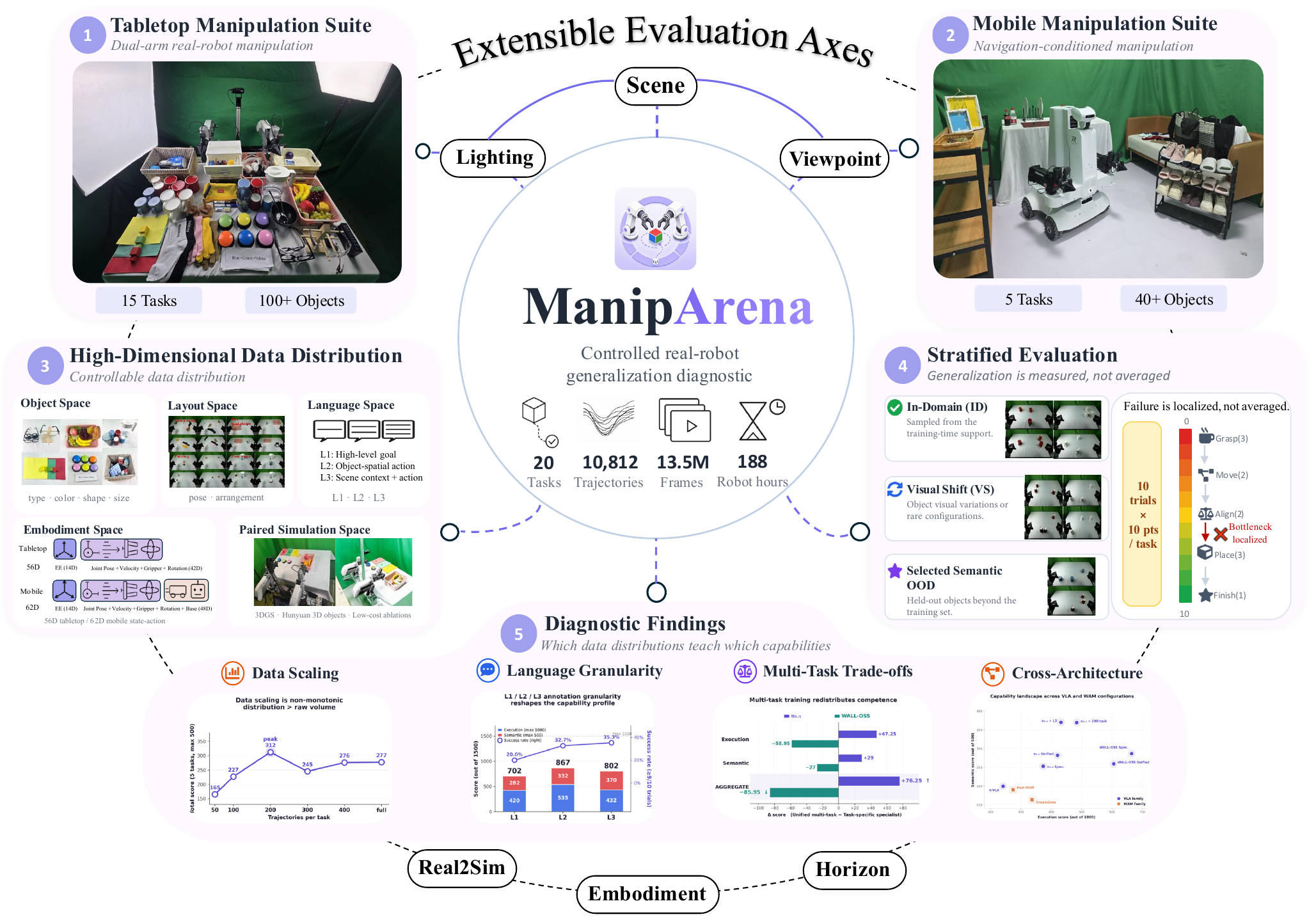}
    \caption{\textbf{ManipArena at a glance.} ManipArena is a controlled
    real-robot generalization diagnostic comprising $20$ tasks,
    $10{,}812$ trajectories, $13.5$M frames, and approximately
    $188$ robot hours. The task suite combines a 15-task
    \textbf{tabletop platform} (panel \textbf{1}) and a 5-task
    \textbf{mobile manipulation platform} (panel \textbf{2}).
    Demonstrations are sampled across a
    \textbf{high-dimensional data distribution} (panel \textbf{3})
    that exposes object, layout, language (L1/L2/L3), embodiment, and
    paired-simulation axes as controllable training-time variables.
    Each task is evaluated under a \textbf{stratified $10$-trial
    protocol} (panel \textbf{4}) covering in-domain, visual-shift, and
    selected semantic-OOD conditions, with subtask partial-credit
    scoring that localizes failures to grasping, transport, alignment,
    placement, or retraction stages rather than averaging them into a
    single success rate. These design choices support a coordinated set
    of \textbf{diagnostic studies} (panel \textbf{5}) on data scaling,
    language granularity, multi-task trade-offs, and cross-architecture
    comparison, together with a paired-simulation transfer extension.}
    \label{fig:teaser}
\end{figure}

\begin{abstract}
Vision-Language-Action (VLA) models and world-action models have emerged as central paradigms for general-purpose robotic intelligence, yet their empirical progress remains constrained by the absence of evaluation protocols that are both physically realistic and diagnostically controlled. Simulator-centric benchmarks provide scale and reproducibility, but cannot fully capture the reality gap induced by perception noise, contact dynamics, latency, calibration error, and hardware constraints. Conversely, real-robot evaluations are often fragmented across platforms, scenes, objects, and scoring rules, making fair comparison and failure attribution difficult. We introduce \textbf{ManipArena}, a standardized real-robot evaluation framework for studying manipulation generalization under matched physical conditions. ManipArena comprises $20$ tasks, $10{,}812$ expert trajectories, $13.5$M frames, and approximately $188$ robot hours across tabletop and mobile manipulation. The framework combines schema-defined task variation, stratified in-domain, visual-shift, and semantic-OOD trials, subtask-level partial-credit scoring, three-level language annotations, low-level motor signals, and paired real-to-sim environments reconstructed from physical scenes. Using ManipArena, we evaluate seven tabletop configurations spanning VLA and world-action-model policies. The results show that real-robot conclusions depend not only on architecture, but also on model provenance, fine-tuning regime, data sampling, and annotation granularity. ManipArena thus provides a reproducible and interpretable foundation for diagnosing capability boundaries and failure modes in embodied generalization.
\end{abstract}

\section{Introduction}
\label{sec:intro}


Recent advances in robot learning have positioned Vision-Language-Action (VLA) models~\cite{intelligence2025pi_,zhai2025ignitingvlmsembodiedspace,brohan2023rt2visionlanguageactionmodelstransfer,octomodelteam2024octoopensourcegeneralistrobot,zheng2025xvlasoftpromptedtransformerscalable,zhang2026a1fullytransparentopensource,wu2026pragmaticvlafoundationmodel} and World-Action-Model (WAM) policies~\cite{ye2026world,yuan2026fastwamworldactionmodels,wu2023unleashinglargescalevideogenerative,cheang2024gr2generativevideolanguageactionmodel,bi2025motusunifiedlatentaction,li2026causalworldmodelingrobot} at the forefront of manipulation research. By unifying vision, language, and action within a single policy interface, these models have demonstrated strong performance across tabletop and mobile manipulation, as well as long-horizon instruction following. As capabilities scale, however, evaluation has emerged as a key bottleneck. A policy may appear general under one robot, scene, or task distribution yet fail under another; conversely, aggregate metrics may obscure competent subskills masked by a few physical bottlenecks. Progress therefore requires evaluation protocols that are not only realistic, but also sufficiently standardized and diagnostic to enable interpretable performance assessment.


The dominant evaluation paradigm in robot learning reflects a persistent tradeoff between reproducibility and physical validity. Existing benchmarks typically optimize for one at the expense of the other: they either provide controlled, large-scale evaluation in simulation, or capture real-world complexity with limited standardization. Simulator-centric benchmarks (e.g., RLBench~\cite{james2020rlbench}, CALVIN~\cite{mees2022calvin}, LIBERO~\cite{liu2023libero}, ManiSkill~\cite{tao2025maniskill3gpuparallelizedrobotics}, VLABench~\cite{zhang2025vlabench}) offer reproducible protocols but rely on simplified dynamics and weakly controlled distribution shifts, making it unclear whether performance reflects genuine reasoning or dataset regularities. In contrast, real-robot evaluations capture realistic factors (\eg, perception noise, contact dynamics, latency), yet are often tied to specific setups, limiting comparability across studies. More broadly, current protocols either focus on short-horizon tasks or consider complex settings without systematic control; they either report aggregate metrics or provide fine-grained diagnostics at the cost of scalability; and they either enable scalable simulation or faithfully capture real-world physics, but rarely both. Consequently, attributing performance to reasoning, data, or task design remains challenging.

To address these challenges, we introduce ManipArena, a real-robot evaluation framework for standardized and attribution-aware assessment. It is built upon five core principles:
\begin{itemize}[topsep=0pt, partopsep=0pt, leftmargin=13pt, parsep=0pt, itemsep=3pt]
    
    
    
    
    \item \textbf{Reasoning-Oriented:} ManipArena prioritizes reasoning-oriented manipulation by incorporating complex spatial constraints, multi-stage bimanual operations, and semantic understanding that go beyond simple motor skills, requiring models to demonstrate genuine execution reasoning or semantic reasoning rather than trajectory memorization.

    \item \textbf{Multi-Level Generalization:} It facilitates controlled multi-level generalization through a green-screen enclosed environment with fixed lighting, systematic diversity guides for data collection, and layered out-of-distribution testing across materials, appearances, and spatial configurations.

    \item \textbf{Mobile Manipulation:} Beyond tabletop tasks, ManipArena incorporates long-horizon mobile manipulation requiring navigation, spatial memory, and sustained whole-body control across an extended workspace, providing comprehensive coverage of real-world operational scenarios.

    \item \textbf{Rich Sensory Diagnostics:} The framework provides rich and open sensory signals, including low-level motor currents and joint velocities beyond standard joint states, allowing developers to perform granular force-aware action modeling and diagnostic analysis.

    \item \textbf{Real-to-Sim Synchronization:} By utilizing high-quality 3D scanning, we provide synchronized real and simulated environments, constructing physically consistent simulation counterparts to enable scalable evaluation and sim-to-real diagnostics without sacrificing physical realism.
\end{itemize}

Specifically, ManipArena comprises 20 real-robot tasks, 10,812 trajectories, and approximately 188 hours of demonstrations. The benchmark includes 15 tabletop tasks and 5 mobile-manipulation tasks. Tabletop tasks span both execution-heavy manipulation (\eg, insertion, pouring, bimanual placement, rearrangement) and semantic manipulation (\eg, classification, ordering, matching, category-based selection), while mobile tasks extend to navigation-conditioned, long-horizon scenarios. Each trajectory is annotated at three language levels: L1 provides abstract task instructions, L2 specifies spatially grounded actions, and L3 incorporates scene-level context. This design treats language granularity as a controllable experimental variable rather than a fixed dataset attribute.

    
    
    
    

Using the ManipArena protocol, we evaluate seven tabletop baseline configurations spanning VLA and WAM policy families, including $\pi_{0.5}$, WALL-OSS-0.5, X-VLA, DreamZero, and Fast-WAM, with both task-specific and multi-task variants. \textbf{Key findings are as follows:}
\begin{itemize}[topsep=0pt, partopsep=0pt, leftmargin=13pt, parsep=0pt, itemsep=3pt]
\item \textbf{Unsaturated benchmark:}
The best aggregate tabletop score (WALL-OSS-0.5-Specialist) reaches 951/1500 (63.4\%), and no policy consistently dominates across tasks.

\item \textbf{Nontrivial effects of task sharing:}
Multi-task training improves $\pi_{0.5}$ (626 $\rightarrow$ 703) but degrades WALL-OSS-0.5 (951 $\rightarrow$ 865), indicating sensitivity to model design and task structure.

\item \textbf{Balance vs.\ scale:}
A balanced setting (200 trajectories per task) outperforms a larger-scale regime ($\sim$500 per task) for $\pi_{0.5}$.

\item \textbf{Task-dependent language granularity:} L2-level tasks yield the strongest execution performance, whereas L3-level tasks are more effective for semantic reasoning.

\item \textbf{Challenges in mobile manipulation:} ManipArena includes five navigation-conditioned mobile tasks that extend evaluation beyond tabletop settings. Current policy evaluation is limited to a scoped task-specific $\pi_{0.5}$ configuration, with per-task results reported in Appendix~\ref{app:per_task}.

\end{itemize}

\begin{table}[t]
\centering
\setlength{\tabcolsep}{10pt}
\renewcommand{\arraystretch}{1.1}
\caption{\textbf{Comparison of ManipArena with existing benchmarks.} Prior works are either simulator-centric or lack standardized real-world evaluation. ManipArena uniquely provides a comprehensive, standardized framework bridging simulation and real-world deployment with reasoning-intensive tasks and multi-level generalization.}
\vspace{1mm}
\resizebox{\linewidth}{!}{
\begin{tabular}{lccccccc}
\toprule
\textbf{Benchmark} 
& \textbf{Env} 
& \textbf{Reasoning} 
& \textbf{Generalization} 
& \textbf{Mobile} 
& \textbf{Sensory} 
& \textbf{Real-to-Sim} \\
\midrule
RLBench~\cite{james2020rlbench} & Sim   & Low & Limited & \ding{55} & \ding{55} & \ding{55} \\
LIBERO~\cite{liu2023libero} & Sim & Low & Moderate & \ding{55} & \ding{55} & \ding{55} \\
CALVIN~\cite{mees2022calvin} & Sim & Medium & Moderate & \ding{55} & \ding{55} & \ding{55} \\
ManiSkill2~\cite{gu2023maniskill2} & Sim  & Low & Strong & \ding{51} & \ding{55} & \ding{55} \\
VLABench~\cite{zhang2025vlabench} & Sim  & \textbf{High} & Strong & \ding{55} & \ding{55} & \ding{55} \\
RoboArena \cite{atreya2025roboarena} & Real  & Medium & Weak & \ding{51} & \ding{55} & \ding{55} \\
RoboChallenge \cite{yakefu2025robochallenge} & Real  & Medium & Partial & \ding{51} & \ding{55} & \ding{55} \\
ManipulationNet \cite{chen2026manipulationnet} & Real & Medium & Weak & \ding{55} & \ding{55} & \ding{51} \\
\textbf{ManipArena (Ours)} & Real  & \textbf{High}  & \textbf{Systematic}  & \ding{51}  & \ding{51}  & \ding{51} \\
\bottomrule
\end{tabular}}
\end{table}

Our contributions are threefold:
\begin{itemize}[topsep=0pt, partopsep=0pt, leftmargin=13pt, parsep=0pt, itemsep=3pt]
\item We design a green-screen-based evaluation setup with fixed hardware and environment configurations, together with schema-driven task construction and stratified protocols, enabling systematic control over task variation and distribution shifts.

\item ManipArena comprises 20 real-robot tasks, 10,812 trajectories, and 188 hours of demonstrations, spanning both tabletop and mobile manipulation. Each trajectory is annotated at three language levels (L1–L3), treating language granularity as a controllable experimental variable.

\item We benchmark representative VLA and WAM policies under a unified protocol and provide detailed insights into task sharing, data scaling, language granularity, and long-horizon manipulation, revealing key limitations of current approaches.
\end{itemize}

\section{Related Work}
\label{sec:related}
\subsection{Benchmarking for Robot Manipulation}

A large body of work has focused on developing benchmarks for robot manipulation to evaluate learning-based policies. Early simulation-centric benchmarks, such as RLBench \cite{james2020rlbench}, provide a diverse set of tasks with scalable demonstration generation and multi-modal observations, enabling reproducible evaluation across a wide range of learning paradigms. Subsequent efforts extend this direction toward more complex manipulation. For example, CALVIN \cite{mees2022calvin} introduces compositional long-horizon tasks grounded in language instructions, emphasizing sequential decision-making and cross-task generalization. LIBERO \cite{liu2023libero} further explores lifelong learning and knowledge transfer, providing structured task suites for evaluating continual learning in embodied agents. More recent benchmarks aim to improve diversity and generalization. ManiSkill2 \cite{gu2023maniskill2} introduces a unified simulation platform with large-scale object diversity, dynamic physics, and multi-task support, significantly improving coverage of manipulation scenarios. VLABench \cite{zhang2025vlabench} further incorporates reasoning-intensive tasks, requiring models to integrate language understanding, spatial reasoning, and world knowledge for long-horizon manipulation.

Despite these advances, most existing benchmarks remain predominantly simulation-based. While simulators offer scalability and controllability, they fail to capture real-world complexities such as perception noise, contact dynamics, and system latency. Although some recent efforts \cite{fei2025liberoplusindepthrobustnessanalysis,zhou2025liberoprorobustfairevaluation,pumacay2024colosseumbenchmarkevaluatinggeneralization,kim2026molmospaceslargescaleopenecosystem} have introduced richer object diversity, dynamic physics they still cannot replicate the full visual, physical, and temporal variation of uncontrolled deployment environments. As a result, performance in simulation often does not reliably translate to real-world deployment, limiting their effectiveness for evaluating general-purpose embodied intelligence.

\subsection{Real-World Evaluation for Robot Learning}

To overcome the limitations of simulation, a growing body of work has explored real-world benchmarks and evaluation frameworks for robot manipulation. Early efforts such as FurnitureBench \cite{heo2025furniturebench} introduces a standardized and reproducible platform for long-horizon manipulation through furniture assembly tasks, supported by large-scale teleoperation data and unified system configurations. SceneReplica \cite{khargonkar2024scenereplica} proposes replicable real-world scenes using standardized objects and layouts, enabling consistent evaluation across different laboratories. Platforms such as the Real Robot Challenge \cite{bauer2022real} provide remote access to standardized robotic systems, allowing policies to be evaluated under consistent hardware and protocols at scale. Recent frameworks including RoboArena \cite{atreya2025roboarena}, RoboChallenge \cite{yakefu2025robochallenge}, and ManipulationNet \cite{chen2026manipulationnet} further extend this direction by enabling distributed real-world benchmarking and attempting to standardize evaluation pipelines across different settings.

Despite these advances, existing real-world evaluation approaches still exhibit several limitations. First, many benchmarks are task-specific or domain-specific, such as furniture assembly or warehouse picking, limiting their generality. Second, most setups lack systematic control over generalization factors, making it difficult to disentangle the effects of appearance, spatial configuration, and task variation. Third, current benchmarks rarely emphasize reasoning-oriented manipulation, instead focusing primarily on execution-level skills such as grasping or pick-and-place. As a result, there remains a critical need for a unified, controlled, and reasoning-oriented real-world evaluation framework that enables fair comparison, interpretable diagnostics, and systematic analysis of generalization in embodied AI systems.

\section{ManipArena Benchmark}                                                                                                                            
\label{sec:benchmark}
ManipArena is a schema-driven real-robot benchmark for evaluating
manipulation generalization under controlled physical conditions. It contains
two robot platforms, 20 tasks, 10,812 demonstrations, a stratified 10-trial
evaluation protocol, partial-credit scoring, three-level language annotations,
and a paired simulation extension. We describe the benchmark components needed
for the main experiments; per-task statistics, observation/action fields,
sampling procedures, rubrics, and OOD assignments are deferred to
Appendix~\ref{app:dataset_stats}--\ref{app:ood_audit}.

\subsection{Controlled setup and benchmark scale}
\label{sec:setup}

\begin{figure}[t]
\centering
\begin{subfigure}[b]{0.48\textwidth}
    \includegraphics[width=\textwidth]{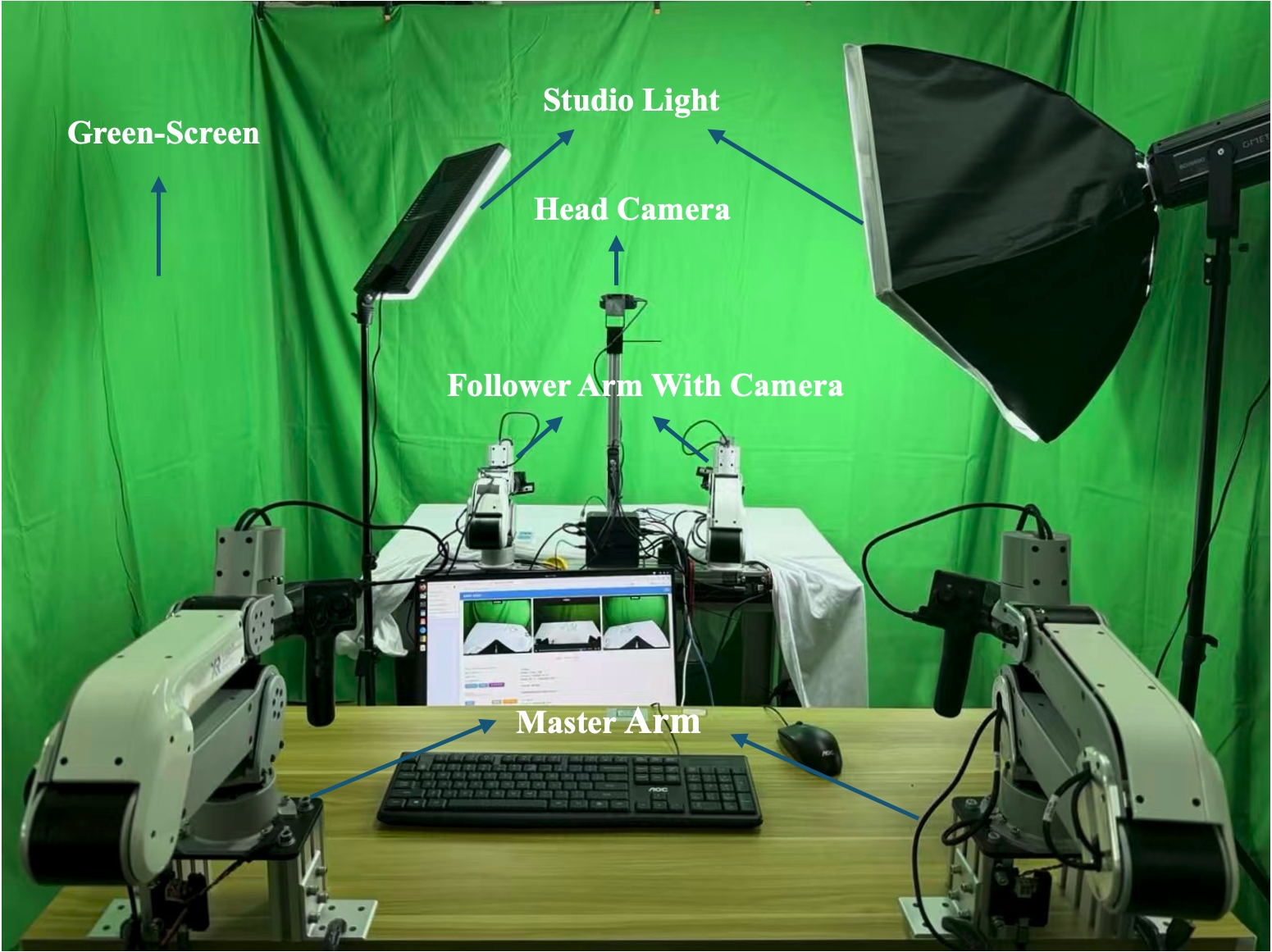}
    \caption{Tabletop platform.}
    \label{fig:setup_tabletop}
\end{subfigure}
\hfill
\begin{subfigure}[b]{0.48\textwidth}
    \includegraphics[width=\textwidth]{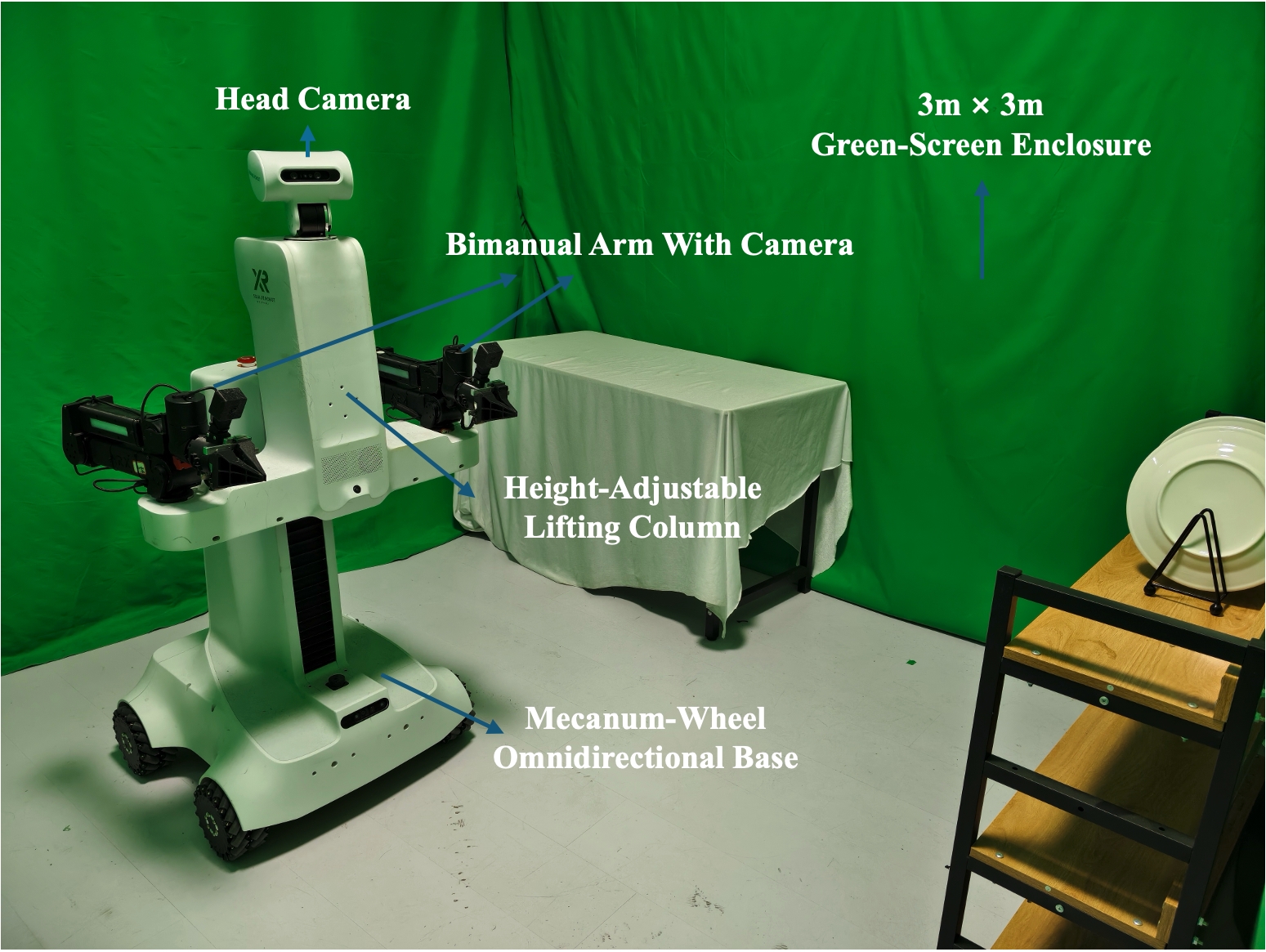}
    \caption{Mobile platform.}
    \label{fig:setup_mobile}
\end{subfigure}
\caption{Physical platforms used by ManipArena. The tabletop platform supports
15 fixed-base bimanual manipulation tasks, while the mobile platform supports
5 longer-horizon navigation-conditioned manipulation tasks. Both are evaluated
inside controlled green-screen environments.}
\label{fig:setup}
\end{figure}

ManipArena uses physical execution as the final measurement. The tabletop
platform is a fixed bimanual system with a head camera and two wrist cameras;
the mobile platform adds a lifting column and omnidirectional base inside a
$3\,\mathrm{m}\times3\,\mathrm{m}$ reconfigurable enclosure
(Figure~\ref{fig:setup}). Both platforms run at 20\,Hz in green-screen booths
with fixed lighting, camera placement, and workspace geometry, and release
proprioceptive state/action vectors (Appendix~\ref{app:data_format}). This
control is used to reduce incidental physical confounds during evaluation,
not to claim that deployment scenes are visually simple.

The benchmark contains $20$ real-robot tasks, $10{,}812$ demonstrations,
$13.5$M frames, and approximately $188$ hours of data. The 15 tabletop tasks
support the cross-model studies in Section~\ref{sec:experiments}; the 5
mobile tasks provide a scoped long-horizon manipulation assessment.
Table~\ref{tab:benchmark_summary} summarizes the task families.

\begin{table}[t]
\centering
\small
\caption{ManipArena task families. Durations are averaged over collected trajectories.}
\label{tab:benchmark_summary}
\setlength{\tabcolsep}{5pt}
\begin{tabular}{@{}lcccl@{}}
\toprule
Family & Tasks & Traj. & Avg. dur. & Evaluation role \\
\midrule
Tabletop execution & 10 & 5,157 & 39.2s & Contact, placement, insertion, pouring \\
Tabletop semantic & 5 & 2,783 & 25.0s & Category, order, matching, object selection \\
Mobile manipulation & 5 & 2,872 & 143.9s & Navigation-conditioned long-horizon manipulation \\
\midrule
\textbf{Total} & \textbf{20} & \textbf{10,812} & -- & \textbf{13.5M frames, 188 hours} \\
\bottomrule
\end{tabular}
\end{table}

\subsection{Schema-driven task design}
\label{sec:schema_diversity}

The central unit of ManipArena is a task schema. A schema specifies the task
stages, object and layout variables, distractor condition, and held-out
condition reserved for evaluation. Demonstrations and evaluation trials are
therefore sampled from declared regions of the same task space, making
train/evaluation overlap and held-out variation explicit.

Formally, for task $i$, a sampled configuration
$c=(d_{i,1},\ldots,d_{i,M_i},\delta)$ contains the values of its diversity
dimensions and distractor condition. The training collection draws
\begin{equation}
\label{eq:schema}
c \sim P_i^{\mathrm{train}},
\end{equation}
where $P_i^{\mathrm{train}}$ is specified by the task schema. The task families
separate execution-heavy manipulation, semantic object reasoning, and
navigation-conditioned mobile manipulation (Table~\ref{tab:benchmark_summary});
full schema definitions are given in Appendix~\ref{app:protocol}.

\subsection{Stratified real-robot evaluation}
\label{sec:stratified_eval}

For each tabletop task, ManipArena uses a fixed 10-trial evaluation protocol.
Trials 1--4 sample in-domain configurations, trials 5--8 introduce controlled
visual or configuration shifts, and trials 9--10 use the strongest held-out
condition available for the task:
\begin{equation}
\label{eq:strat}
c_t \sim
\begin{cases}
P_i^{\mathrm{ID}}, & t \in \{1,\ldots,4\}, \\
P_i^{\mathrm{shift}}, & t \in \{5,\ldots,8\}, \\
P_i^{\mathrm{held}}, & t \in \{9,10\},
\end{cases}
\end{equation}
where all three distributions are induced by the same task schema. The final
band is task-dependent: 7 tabletop tasks use object-level OOD conditions,
while the remaining 8 test spatial, ordering, or procedural held-out
conditions. The per-task audit is provided in Appendix~\ref{app:ood_audit}.

\subsection{Partial-credit scoring}
\label{sec:subtask_scoring}

ManipArena scores each trial with task-specific subgoals rather than a binary
success label. Each trial is worth 10 points, distributed over stages such as
grasping, transport, alignment, insertion, pouring, semantic selection, and
final retraction:
\begin{equation}
\label{eq:score}
S_i^{(t)}
= \sum_{k=1}^{K_i} w_{i,k}\,
\mathbf{1}\!\left[g_{i,k}\text{ achieved on trial }t\right],
\qquad
\sum_{k=1}^{K_i} w_{i,k}=10.
\end{equation}
For a task subset $\mathcal{I}$, we report aggregate partial-credit score and
success rate, where success requires at least $9/10$ points:
\begin{equation}
\label{eq:agg_sr}
S(\mathcal{I})=\sum_{i\in\mathcal{I}}\sum_{t=1}^{10} S_i^{(t)}, \qquad
\mathrm{SR}(\mathcal{I}) =
\frac{1}{|\mathcal{I}|}\sum_{i\in\mathcal{I}}\frac{1}{10}
\sum_{t=1}^{10}\mathbf{1}\!\left[S_i^{(t)}\geq 9\right].
\end{equation}

Partial credit exposes whether a policy fails at perception, grasping,
alignment, semantic selection, or final completion, which binary success would
collapse. Per-band aggregation and evaluation algorithms are provided in
Appendix~\ref{app:protocol}; per-task rubrics are summarized in
Appendix~\ref{app:rubrics}.

\subsection{Language annotations and simulation extension}
\label{sec:language_annotations}
\label{sec:paired_sim}

Each trajectory includes three language fields: L1 is a task-level
instruction, L2 adds episode-specific object and spatial information, and L3
adds a generated scene description. This makes language granularity a
controlled training variable in Section~\ref{sec:language}.

ManipArena also includes paired simulation environments for three tabletop
tasks. They share task schemas and observation/action conventions with the
real setup and are used only as a scoped screening diagnostic in
Section~\ref{sec:real2sim}, before real-robot validation.
                                                                                                                                                                         
\section{Experiments}
\label{sec:experiments}                                                                                                                   

We use ManipArena as a controlled diagnostic for real-robot generalization, not only as a leaderboard. Across the 15 tabletop tasks, we evaluate seven baseline configurations spanning VLA and world-action-model (WAM) policies, task-specific and unified fine-tuning, and distinct pretraining provenance. Mobile manipulation is included as a task-suite extension, with scoped task-specific $\pi_{0.5}$ results on the five mobile tasks reported in Appendix~\ref{app:per_task}. The studies below vary one evaluative axis at a time---data quantity, language annotation, training regime, model provenance, or real/sim source---while using the stratified trial protocol and partial-credit scoring of Sections~\ref{sec:stratified_eval}--\ref{sec:subtask_scoring}. Within each model family and controlled study, we keep the implementation family, observation/action interface, test prompt convention, and evaluation protocol fixed unless that factor is explicitly varied; details are summarized in Appendix~\ref{app:baseline_config}.

\subsection{Data scaling and sampling effects}
\label{sec:data_scaling}

We first test whether increasing demonstration count produces monotonic real-robot gains. Holding the architecture, unified training regime, language field, robot platform, and scoring rule fixed, we fine-tune $\pi_{0.5}$ on 50, 100, 200, 300, and 400 trajectories per task, and on the full dataset. All runs use 60k steps except a full-data 120k-step control. Full per-task results are given in Appendix~\ref{app:data_scaling_detail}.

\begin{figure}[t]
    \centering
    \begin{minipage}[t]{0.38\linewidth}
        \vspace{0pt}
        \centering
        \setlength{\tabcolsep}{10pt}
        \renewcommand{\arraystretch}{1.08}
        \begin{tabular}{@{}lccc@{}}
            \toprule
            Data & Steps & Sc. & SR \\
            \midrule
            50/task & 60k & 165 & 10 \\
            100/task & 60k & 227 & 16 \\
            \textbf{200/task} & 60k & \textbf{312} & \textbf{42} \\
            300/task & 60k & 245 & 30 \\
            400/task & 60k & 276 & 36 \\
            Full & 60k & 277 & 34 \\
            Full\textsuperscript{\dag} & 120k & 248 & 26 \\
            \bottomrule
        \end{tabular}
    \end{minipage}
    \hfill
    \begin{minipage}[t]{0.56\linewidth}
        \vspace{0pt}
        \centering
        \includegraphics[width=\linewidth]{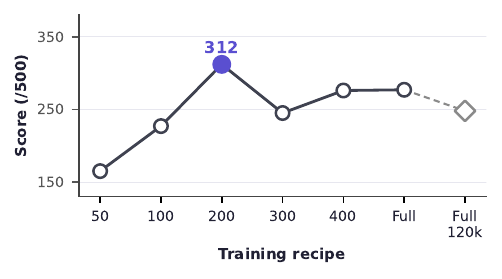}
    \end{minipage}
    \caption{Data scaling for the unified $\pi_{0.5}$ policy on five representative tasks. The table reports five-task totals (score out of 500; SR in \%), and the curve visualizes the same trend. The 200/task subset obtains the highest total score, while larger subsets, the full-data recipe, and the full-data 120k-step control do not improve monotonically. \textsuperscript{\dag}Full-data model trained for 120k steps; all other rows use 60k steps. Per-task results are given in Appendix~\ref{app:data_scaling_detail}.}
    \label{fig:data_scaling}
\end{figure}

The aggregate trend is non-monotonic (Figure~\ref{fig:data_scaling}). The 200/task recipe obtains the best five-task score (312/500, 42\% SR), exceeding the full-data 60k recipe (277/500, 34\% SR) and the full-data 120k control (248/500, 26\% SR). The same ordering appears in the full 15-task evaluation: 200/task reaches 853/1500 and 35.3\% SR, while full-data 120k reaches 727.3/1500 and 23.3\% SR. This result should not be read as a universal optimum at 200 demonstrations per task; rather, it shows that raw demonstration count is an incomplete proxy for policy quality, and that sample balance and task composition can materially affect real-robot conclusions.

\subsection{Language annotation granularity}
\label{sec:language}

\begin{wraptable}{r}{0.45\linewidth}
\vspace{-10pt}
\centering
\small
\setlength{\tabcolsep}{5pt}
\renewcommand{\arraystretch}{1.12}
\caption{Effect of unified multi-task fine-tuning relative to task-specific specialists on the 15 tabletop tasks. Scores are category totals; positive values indicate gains under unified training.}
\label{tab:multitask_summary}
\begin{tabular}{lccc}
\toprule
Model & $\Delta$ Exec & $\Delta$ Sem & $\Delta$ Total \\
\midrule
$\pi_{0.5}$ & +47.3 & +29.0 & +76.3 \\
WALL-OSS-0.5 & -59.0 & -27.0 & -86.0 \\
\bottomrule
\end{tabular}
\vspace{-10pt}
\end{wraptable}

We next vary the language field paired with each demonstration while evaluating all policies with the same L1 task-level prompt. The three unified $\pi_{0.5}$ policies differ only in training-time annotation: L1 is a task goal, L2 adds episode-specific object and spatial information, and L3 further adds a generated scene description.

\begin{table}[t]
\centering
\small
\vspace{1mm}
\caption{Language annotation granularity ($\pi_{0.5}$, unified). L1 is a task-level high-level goal, L2 is a per-episode scene-specific instruction, and L3 adds the generated scene description to the cleaned L2-style instruction. All models are evaluated with L1 instructions. SR in \%.}
\label{tab:language}
\setlength{\tabcolsep}{12pt}
\renewcommand{\arraystretch}{1.2}
\definecolor{l2col}{RGB}{217, 232, 252}
\definecolor{l3col}{RGB}{252, 228, 236}
\begin{tabular}{l cc
>{\columncolor{l2col}[8pt][\tabcolsep]}c
>{\columncolor{l2col}[\tabcolsep][10pt]}c
>{\columncolor{l3col}[10pt][\tabcolsep]}c
>{\columncolor{l3col}[\tabcolsep][8pt]}c
}
\toprule
& \multicolumn{2}{c}{L1 abstract}
& \multicolumn{2}{>{\columncolor{l2col}[8pt][10pt]}c}{\textbf{L2 grounded}}
& \multicolumn{2}{>{\columncolor{l3col}[10pt][8pt]}c}{\textbf{L3 scene+grounded}} \\
\cmidrule(lr){2-3} \cmidrule(lr){4-5} \cmidrule(lr){6-7}
Category & Sc & SR & Sc & SR & Sc & SR \\
\midrule
Execution\,(/1000) & 421 & 15 & \textbf{535} & \textbf{28} & 432 & 25 \\
Semantic\,(/500)   & 282 & 28 & 333 & 42 & \textbf{370} & \textbf{56} \\
\midrule
\textbf{Total}\,(/1500) & 703 & 20 & \textbf{867} & 33 & 802 & \textbf{35} \\
\bottomrule
\end{tabular}
\end{table}

\begin{figure}[t]
    \centering
    \includegraphics[width=\linewidth]{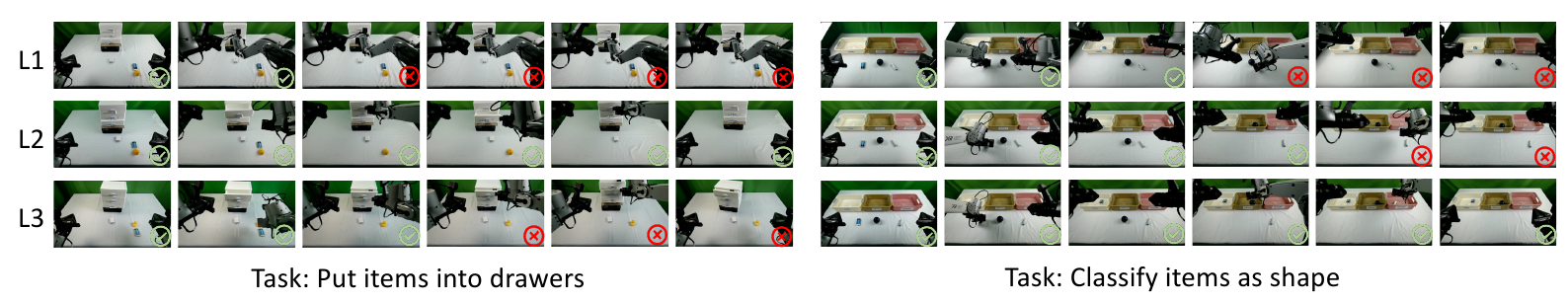}
    \caption{Qualitative comparison of language annotation granularity. Left:
    for an execution-heavy drawer task, L2 provides object- and
    location-grounded information that supports the manipulation sequence,
    whereas L1 lacks scene grounding and L3 adds broader context that is less
    directly action-centered. Right: for a semantic classification task, the
    additional scene context in L3 improves object-level disambiguation.}
    \label{fig:language_qualitative}
\end{figure}

Scene-specific language improves performance despite being absent at test time (Table~\ref{tab:language}). L2 raises total score from 703 to 867 and SR from 20\% to 32.7\%, while L3 obtains the highest SR (35.3\%) and strongest semantic score (370/500). The category split is informative: L2 is best on execution tasks (535/1000), whereas L3 shifts performance toward semantic reasoning while reducing execution score to 432/1000. Figure~\ref{fig:language_qualitative} illustrates the same distinction qualitatively. Annotation granularity should therefore be treated as part of the training distribution: action-centered grounding benefits spatial manipulation, whereas broader scene context is more useful for semantic disambiguation. Per-task results are given in Appendix~\ref{app:language_detail}.

\subsection{Task-level trade-offs in multi-task training}
\label{sec:single_multi}

We next isolate the fine-tuning regime by comparing task-specific specialists with a unified multi-task checkpoint for both $\pi_{0.5}$ and the WALL-OSS-0.5 reference model. The robot platform, task set, trial protocol, and scoring rule are fixed.



The effect has opposite signs across architectures (Table~\ref{tab:multitask_summary}). Unified training improves $\pi_{0.5}$ by 76.3 points, but reduces WALL-OSS-0.5 by 86.0 points. Per-task results in Appendix~\ref{app:per_task} show that these shifts are competence redistributions rather than uniform gains or losses. The pattern is consistent with a provenance-based interpretation: WALL-OSS-0.5 is pretrained on data from a closely related embodiment, whereas $\pi_{0.5}$ uses broader cross-platform pretraining. The evidence is correlational, but it shows that benchmark rankings depend on the interaction between architecture, pretraining distribution, and fine-tuning regime.

\subsection{Cross-architecture evaluation}
\label{sec:model_zoo}

We then place these controlled studies in a broader model landscape. Under the same tabletop protocol, we compare seven configurations spanning VLA and WAM policies: task-specific and unified variants of $\pi_{0.5}$~\cite{intelligence2025pi_} and the embodiment-aligned WALL-OSS-0.5 reference model, together with X-VLA, DreamZero~\cite{ye2026world}, and Fast-WAM. WALL-OSS-0.5 naming and release scope are clarified in Appendix~\ref{app:model_details}. Figure~\ref{fig:total_scores} includes the $\pi_{0.5}$ diagnostic variants from Sections~\ref{sec:data_scaling}--\ref{sec:language} to expose recipe effects alongside model-family differences.

\begin{figure}[t]
    \centering
    \begin{subfigure}[t]{0.48\linewidth}
        \centering
        \includegraphics[width=\linewidth]{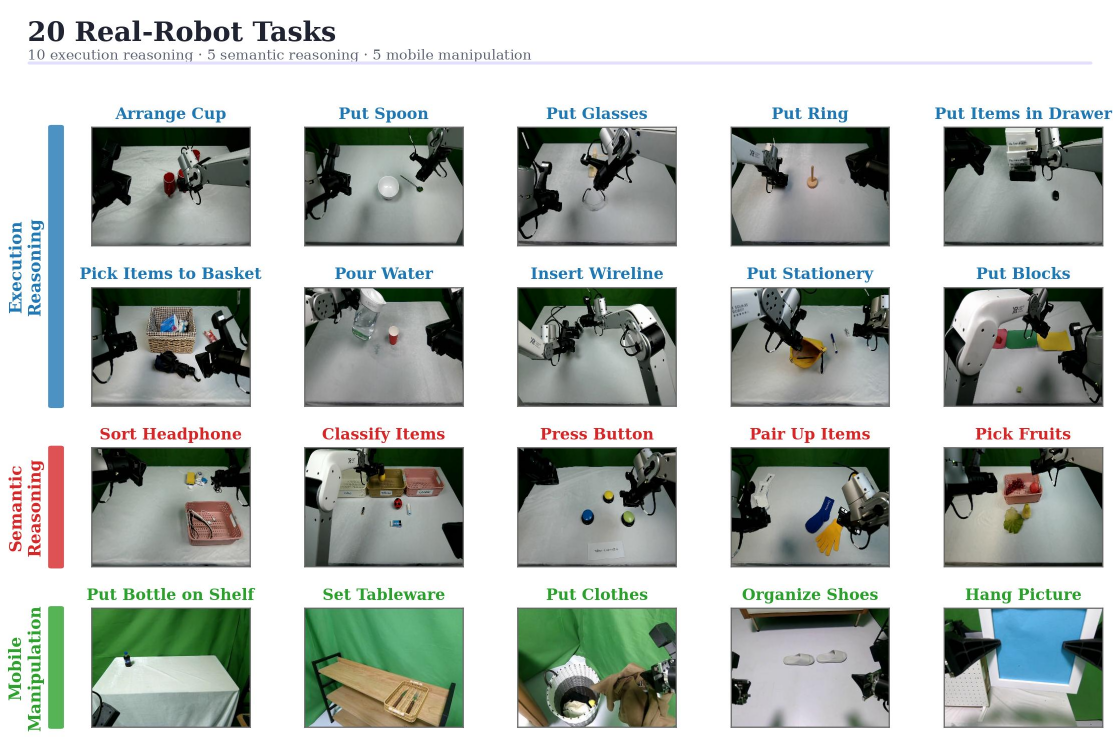}
        \caption{Task coverage.}
        \label{fig:task_overview_grid}
    \end{subfigure}
    \hfill
    \begin{subfigure}[t]{0.5\linewidth}
        \centering
        \includegraphics[width=\linewidth]{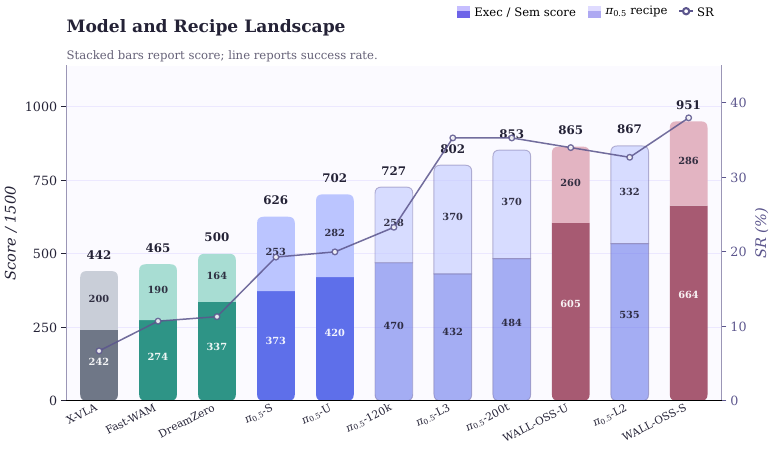}
        \caption{Score and success-rate landscape.}
        \label{fig:score_landscape}
    \end{subfigure}
    \caption{ManipArena task suite and tabletop evaluation landscape. (a) The benchmark spans 10 tabletop execution tasks, 5 tabletop semantic tasks, and 5 mobile manipulation tasks. (b) Tabletop-only aggregate scores for the 15-task evaluation. Stacked bars decompose execution and semantic scores, while the line reports mean success rate; lighter $\pi_{0.5}$ bars denote recipe variants from the data-scaling and language experiments.}
    \label{fig:total_scores}
\end{figure}

\begin{table}[t]
\centering
\caption{Model zoo on 15 tabletop tasks. Scores are category totals (Execution out of 1000, Semantic out of 500). SR is mean success rate (\%). Spec.\ = task-specific fine-tuning; Unified = unified multi-task. Full per-task results in Appendix~\ref{app:per_task}.}
\label{tab:model_zoo}
\setlength{\tabcolsep}{8pt}
\renewcommand{\arraystretch}{1.1}
\scriptsize
\resizebox{\linewidth}{!}{%
\begin{tabular}{@{}l cc cc cc cc cc cc cc cc@{}}
\toprule
& \multicolumn{4}{c}{$\pi_{0.5}$\cite{intelligence2025pi_}}
& \multicolumn{4}{c}{\textbf{WALL-OSS-0.5 \cite{zhai2025ignitingvlmsembodiedspace}}}
& \multicolumn{2}{c}{X-VLA\cite{zheng2025xvlasoftpromptedtransformerscalable}}
& \multicolumn{2}{c}{Fast-WAM\cite{yuan2026fastwamworldactionmodels}}
& \multicolumn{2}{c}{DreamZero\cite{ye2026world}}
& \multicolumn{2}{c}{Wall-WM \cite{wallwm2026}} \\
\cmidrule(lr){2-5} \cmidrule(lr){6-9} \cmidrule(lr){10-11} \cmidrule(lr){12-13} \cmidrule(lr){14-15} \cmidrule(lr){16-17}
& \multicolumn{2}{c}{Spec.} & \multicolumn{2}{c}{Unified}
& \multicolumn{2}{c}{Spec.} & \multicolumn{2}{c}{Unified} & \multicolumn{2}{c}{Unified} & \multicolumn{2}{c}{Unified} & \multicolumn{2}{c}{Unified} & \multicolumn{2}{c}{Unified} \\
\cmidrule(lr){2-3} \cmidrule(lr){4-5} \cmidrule(lr){6-7} \cmidrule(lr){8-9} \cmidrule(lr){10-11} \cmidrule(lr){12-13} \cmidrule(lr){14-15} \cmidrule(lr){16-17}
& Sc & SR & Sc & SR & Sc & SR & Sc & SR & Sc & SR & Sc & SR & Sc & SR & Sc & SR \\
\midrule
Exec\,(/1000) & 373 & 15 & 421 & 16 & \textbf{664} & \textbf{44} & 605 & 35 & 242 & 4 & 274 & 9 & 337 & 11 & 570.5 & 30 \\
Sem\,(/500)   & 253 & 28 & 282 & 28 & \textbf{287} & 26 & 260 & \textbf{32} & 200 & 12 & 191 & 14 & 164 & 12 & 297.5  & 34 \\
\midrule
\textbf{Total}\,(/1500) & 626 & 19 & 703 & 20 & \textbf{951} & \textbf{38} & 865 & 34 & 442 & 7 & 465 & 11 & 500 & 11 & 868 & 31.33 \\
\bottomrule
\end{tabular}
}
\end{table}


The task-specific WALL-OSS-0.5 achieves the highest aggregate score (950.8/1500, 38\% SR), with gains primarily concentrated in execution tasks (664.3 vs.\ 420.5 for $\pi_{0.5}$ Unified), while semantic performance is comparable (286.5 vs.\ 282). This suggests that the advantage is better attributed to embodiment-aligned pretraining than to architecture alone. Diagnostic variants of $\pi_{0.5}$ further show that training recipes can rival model-family differences: without architectural changes, performance increases from 702.5 (full-data L1) to 853.0 (200/task) and 867.3 (L2 annotations). Despite these gains, the benchmark remains unsaturated, with the best model reaching 63.4\% of the maximum tabletop score. Overall, real-robot performance reflects a combination of architecture, pretraining, fine-tuning, data composition, and annotation design. ManipArena makes these factors sufficiently disentangled for analysis, rather than collapsing them into a single leaderboard.

\subsection{Paired simulation diagnostics}
\label{sec:real2sim}

As a final diagnostic axis, we evaluate the paired simulation extension of Section~\ref{sec:paired_sim}. The claim is scoped: simulation is used to compare data-source and mixture recipes, while real-robot trials remain the endpoint for capability assessment. We evaluate $\pi_{0.5}$ on three paired tasks with real-data, simulated-data, and mixed real/sim checkpoints.

Real-robot transfer results are reported in Appendix~\ref{app:real2sim_detail}. The effect is task-dependent: simulated-data training improves \texttt{press\_button}, cotraining does not uniformly dominate, and real-data training remains strongest on \texttt{pick\_fruits}. We therefore treat simulation as a screening instrument, not a substitute benchmark: simulated gains require physical validation before being interpreted as real-robot capability gains.

\section{Conclusion}
\label{sec:discussion}


We presented \textbf{ManipArena}, a standardized real-robot evaluation framework for attribution-aware assessment of manipulation policies. By integrating controlled physical environments, schema-driven task design, and stratified protocols, ManipArena enables systematic analysis of generalization, reasoning, and execution under real-world conditions. The benchmark spans diverse tabletop and mobile tasks, with multi-level language annotations that treat language granularity as a controllable experimental variable. Our results show that current VLA and WAM policies remain far from saturated and are highly sensitive to task sharing, data balance, and language specification. These findings expose limitations of existing evaluation practices and motivate more controlled and diagnostic real-world benchmarks. We expect ManipArena to support more rigorous and interpretable evaluation, and to advance research in generalization, long-horizon reasoning, and real-world deployment.

\paragraph{Limitations.} The present conclusions have bounded scope. Real-robot execution is stochastic, and stronger confidence would require more repeated trials. Coverage is constrained by cost: the tabletop study includes seven models, mobile policy evaluation is limited to a scoped task-specific $\pi_{0.5}$ configuration, and the paired simulation extension covers three tasks. Results are also limited to the tested embodiments, controlled environments, and shared L1 instruction interface. Additional details are provided in Appendix~\ref{app:limitations}. These limits define natural extension axes: because ManipArena specifies task schemas, protocols, scoring, language annotations, and simulation assets, future work can vary scenes, embodiments, horizons, and instruction interfaces without redefining the benchmark. ManipArena is therefore intended as a controlled testbed, not a fixed leaderboard.



\bibliography{main.bib}
\bibliographystyle{plain}

\newpage                                
\appendix

\section{Broader Impacts}
\label{broader_impact}
ManipArena can accelerate progress toward more reliable embodied AI agents by
making real-robot generalization measurable and interpretable. However, strong
performance on the benchmark alone should not be assumed sufficient for
real-world deployment. The controlled green-screen environment, fixed lighting,
and standardized object sets do not capture the full visual diversity, clutter,
and dynamic human presence of uncontrolled homes, offices, or factories. Care
should always be exercised when transferring policies evaluated under these
conditions to safety-critical applications involving physical human-robot
interaction or high-stakes manipulation tasks.

\section{Baseline model naming and scope}
\label{app:model_details}


For all model-zoo comparisons, baselines are fine-tuned using the ManipArena
training demonstrations specified for the corresponding setting
(task-specific specialist or unified multi-task) and evaluated with the same
10-trial protocol. No baseline receives additional ManipArena task
demonstrations, evaluation examples, or task-specific real-robot data beyond
the stated training split. This does not make upstream pretraining identical;
differences in pretraining provenance are treated as part of the evaluation.

\section{Baseline implementation consistency}
\label{app:baseline_config}

For each model family, we use a single implementation and configuration family
throughout the corresponding comparisons. The purpose is to ensure that
differences reported in the main text are attributable to the intended
experimental variable rather than to incidental changes in preprocessing,
action representation, prompting, or evaluation. Within each controlled study,
we keep the robot platform, observation streams, action interface,
normalization assets, evaluation prompts, trial protocol, and partial-credit
rubric fixed unless that factor is explicitly the subject of the study.

\paragraph{Controlled changes.}
Task-specific versus unified training changes only the training data grouping
while keeping the model family, action interface, test prompts, and evaluation
protocol fixed. The data-scaling study changes only the number of sampled
demonstrations per task. The language study changes only the training-time
language field, while all policies are evaluated with the same L1 task-level
instruction. The Real2Sim diagnostic changes only the data source or mixture
used for fine-tuning, followed by the same real-robot evaluation endpoint.

\begin{table}[h]
\centering
\small
\caption{Implementation consistency across baseline studies. The table records
which implementation choices are kept fixed within each model family; the main
experiments then vary only the intended diagnostic axis. Mobile policy
evaluation is currently limited to one task-specific $\pi_{0.5}$ configuration
on the five mobile tasks.}
\label{tab:implementation_consistency}
\setlength{\tabcolsep}{4pt}
\begin{tabular}{@{}p{0.16\linewidth}p{0.25\linewidth}p{0.50\linewidth}@{}}
\toprule
Model family & Runs covered & Fixed implementation choices \\
\midrule
$\pi_{0.5}$ / OpenPI &
Task-specific, unified, data scaling, language, scoped mobile, Real2Sim &
OpenPI policy implementation, LeRobot data loaders, normalization assets,
action representation for each platform, and L1 test-time prompting. The
task-specific/unified study changes only task grouping; data scaling changes
only sampled demonstration count; language experiments change only the
training-time language field. \\
\midrule
Fast-WAM &
Tabletop model-zoo baseline &
X2Robot native data adapter, Fast-WAM joint-policy configuration, video
preprocessing, text-embedding pipeline, and tabletop action/proprioceptive
interface are held fixed across the evaluated tabletop tasks. \\
\midrule
DreamZero &
Tabletop WAM baseline and language variants &
XBench 14-DOF data/config family, three-view video input convention, relative
end-effector action representation, DreamZero policy head, and real-robot
inference interface are held fixed. Language variants use matched L1/L2/L3
dataset versions while preserving the same evaluation prompt convention. \\
\bottomrule
\end{tabular}
\end{table}

\section{Experimental resources}
\label{app:resources}

We report the physical and compute resources used for the ManipArena study to
make the evaluation cost and reproducibility assumptions explicit. These
numbers describe data collection, real-robot evaluation, and fine-tuning for
this paper; they exclude upstream pretraining of external foundation models.
The released dataset contains $10{,}812$ trajectories, $13.5$M frames, and
approximately $188$ robot hours collected on the tabletop and mobile robot
platforms in controlled green-screen environments. Beyond robot execution
time, data collection and evaluation require teleoperation, scene reset,
calibration checks, failure recovery, safety supervision, and rubric-based
scoring.

Each tabletop evaluation uses 10 real-robot trials per task. The model-zoo
study alone covers seven configurations across 15 tabletop tasks, corresponding
to at least $7 \times 15 \times 10 = 1{,}050$ tabletop robot trials, before
including the data-scaling, language, Real2Sim, and scoped mobile diagnostic
run.
Table~\ref{tab:training_compute} summarizes the approximate fine-tuning
compute. GPU-hours are computed as the number of GPUs multiplied by wall-clock
training time.

\begin{table}[h]
\centering
\small
\caption{Approximate fine-tuning compute for baseline checkpoints. GPU-hours
exclude upstream pretraining and are intended as resource accounting rather
than a hardware-normalized efficiency comparison.}
\label{tab:training_compute}
\setlength{\tabcolsep}{4pt}
\begin{tabular}{@{}lcccl@{}}
\toprule
Model / setting & Steps & Hardware & Time & GPU-hours \\
\midrule
$\pi_{0.5}$ task-specific & 30k & 8$\times$A800 & 6h & 48 / checkpoint \\
$\pi_{0.5}$ unified multi-task & 60k & 8$\times$A800 & 20h & 160 / checkpoint \\
WALL-OSS-0.5 fine-tuning & -- & 8$\times$A800 & 24h & 192 / checkpoint \\
DreamZero / Fast-WAM & -- & 16$\times$A800 & 72h & 1,152 / checkpoint \\
\bottomrule
\end{tabular}
\end{table}

For reference, 15 tabletop $\pi_{0.5}$ task-specific specialists correspond to
approximately $15 \times 48 = 720$ GPU-hours. Each 60k-step unified
$\pi_{0.5}$ diagnostic run costs approximately 160 GPU-hours, and a 120k-step
control run is approximately 320 GPU-hours under linear scaling. The total
training cost of the paper is therefore larger than a single model-zoo row,
because the data-scaling and language studies train multiple additional
diagnostic checkpoints.

\section{Per-task model zoo results}
\label{app:per_task}
Table~\ref{tab:per_task} reports the complete per-task scores and success rates for all seven baseline configurations.

\begin{table*}[h]
\centering
\small
\caption{Per-task scores and success rates. Score: out of 100 (10 trials $\times$ 10 pts). SR: success rate (\%, $\geq$9/trial). Spec. denotes task-specific specialist fine-tuning and Unified denotes unified multi-task fine-tuning. Best score per task in \textbf{bold}. Mobile policy evaluation is currently limited to the task-specific $\pi_{0.5}$ configuration; unavailable evaluations are marked ``--''.}
\label{tab:per_task}
\setlength{\tabcolsep}{3pt}
\begin{tabular}{@{}cl|cc|cc|cc|cc|cc|cc|cc|cc@{}}
\toprule
& & \multicolumn{4}{c|}{$\pi_{0.5}$} & \multicolumn{4}{c|}{WALL-OSS-0.5} & \multicolumn{2}{c|}{} & \multicolumn{2}{c|}{} & \multicolumn{2}{c}{} & \multicolumn{2}{c}{} \\
& & \multicolumn{2}{c|}{Spec.} & \multicolumn{2}{c|}{Unified} & \multicolumn{2}{c|}{Spec.} & \multicolumn{2}{c|}{Unified} & \multicolumn{2}{c|}{X-VLA} & \multicolumn{2}{c|}{Fast-WAM} & \multicolumn{2}{c}{DreamZero} & \multicolumn{2}{c}{Wall-WM} \\
\# & Task & Sc & SR & Sc & SR & Sc & SR & Sc & SR & Sc & SR & Sc & SR & Sc & SR  & Sc & SR \\
\midrule
\multicolumn{16}{l}{\textit{Execution Reasoning}} \\
\midrule
1 & arrange\_cup & 18 & 0 & 47 & 10 & \textbf{91} & 80 & 58 & 20 & 21 & 0 & 55 & 20 & 25 & 0 & 48 & 0 \\
2 & put\_spoon & 52 & 20 & 32 & 0 & \textbf{80} & 40 & \textbf{80} & 70 & 38 & 20 & 18 & 10 & 54 & 20 & 82 & 70 \\
3 & put\_glasses & 80 & 60 & 67 & 50 & \textbf{90} & 80 & 66 & 60 & 14 & 0 & 38 & 10 & 37 & 0 & 63 & 30 \\
4 & put\_ring & 41 & 20 & 74 & 0 & \textbf{97} & 90 & 91 & 70 & 59 & 10 & 47 & 0 & 27 & 0 & 77 & 40 \\
5 & put\_items\_drw & 34 & 0 & 1 & 0 & 36 & 0 & \textbf{52} & 10 & 16 & 0 & 0 & 0 & 7 & 0 & 67 & 40 \\
6 & put\_blocks & 48 & 30 & 83 & 70 & 95 & 80 & \textbf{96} & 90 & 46 & 10 & 45 & 20 & 27 & 0 & 90 & 80 \\
7 & pour\_water & 8 & 0 & 19 & 0 & \textbf{21} & 0 & 19 & 0 & 4 & 0 & 5 & 0 & 12 & 0 & 21 & 0 \\
8 & insert\_wire & 24 & 0 & 18 & 0 & 30 & 0 & \textbf{50} & 0 & 15 & 0 & 21 & 0 & 24 & 0 & 42 & 0 \\
9 & pick\_items & 47 & 20 & 68 & 30 & 90 & 70 & 75 & 30 & 28 & 0 & 41 & 30 & \textbf{98} & 90 & 60 & 40 \\
10 & put\_stationery & 22 & 0 & 12 & 0 & \textbf{34} & 0 & 19 & 0 & 1 & 0 & 4 & 0 & 26 & 0 & 20.5 & 0 \\
\midrule
\multicolumn{16}{l}{\textit{Semantic Reasoning}} \\
\midrule
11 & sort\_headphone & 35 & 20 & 41 & 20 & 70 & 70 & \textbf{82} & 80 & 36 & 20 & 32 & 0 & 66 & 60 & 84 & 60 \\
12 & classify\_items & \textbf{59} & 20 & 55 & 10 & 55 & 0 & 31 & 0 & 40 & 0 & 25 & 0 & 36 & 0 & 82 & 50 \\
13 & press\_button & \textbf{48} & 20 & 31 & 0 & 22 & 0 & 16 & 0 & 35 & 20 & 0 & 0 & 3 & 0 & 18 & 0\\
14 & pair\_up\_items & 21 & 0 & \textbf{77} & 60 & 68 & 20 & 45 & 10 & 63 & 10 & 63 & 30 & 14 & 0 & 43.5 & 20 \\
15 & pick\_fruits & \textbf{90} & 80 & 78 & 50 & 72 & 40 & 86 & 70 & 26 & 10 & 71 & 40 & 45 & 0 & 70 & 40 \\
\midrule
\multicolumn{16}{l}{\textit{Mobile Manipulation (scoped task-specific $\pi_{0.5}$ run)}} \\
\midrule
16 & put\_clothes & 4.5 & 0 & -- & -- & -- & -- & -- & -- & -- & -- & -- & -- & -- & -- & -- & --\\
17 & hang\_picture & 0 & 0 & -- & -- & -- & -- & -- & -- & -- & -- & -- & -- & -- & -- & -- & --\\
18 & organize\_shoes & 8 & 0 & -- & -- & -- & -- & -- & -- & -- & -- & -- & -- & -- & -- & -- & --\\
19 & put\_bottle & 14 & 0 & -- & -- & -- & -- & -- & -- & -- & -- & -- & -- & -- & -- & -- & --\\
20 & set\_tableware & 10 & 0 & -- & -- & -- & -- & -- & -- & -- & -- & -- & -- & -- & -- & -- & -- \\
\bottomrule
\end{tabular}
\end{table*}

\section{Language Experiment Details}
\label{app:lang_exp_detail}
\label{app:language_detail}

Table~\ref{tab:language_detail} reports the per-task breakdown of the
language annotation experiment. L2
dominates execution-heavy tasks such as \texttt{put\_blocks} and
\texttt{pick\_items}, while L3 achieves the strongest scores on semantic tasks
such as \texttt{classify\_items} and \texttt{pick\_fruits}.

\begin{table*}[t]
\centering
\small
\caption{Language annotation granularity: per-task scores and success rates for $\pi_{0.5}$. Score: out of 100 (10 trials $\times$ 10 pts). SR: success rate (\%, $\geq$9/trial). Best score per task in \textbf{bold}.}
\label{tab:language_detail}
\setlength{\tabcolsep}{4pt}
\begin{tabular}{@{}cl|cc|cc|cc@{}}
\toprule
& & \multicolumn{2}{c|}{L1 Abstract} & \multicolumn{2}{c|}{L2 Grounded} & \multicolumn{2}{c}{L3 Scene+Grounded} \\
\# & Task & Sc. & SR & Sc. & SR & Sc. & SR \\
\midrule
\multicolumn{8}{l}{\textit{Execution Reasoning}} \\
\midrule
1 & arrange\_cup & 18 & 0 & 47 & 10 & \textbf{52} & 30 \\
2 & put\_spoon & 32 & 0 & \textbf{64} & 30 & \textbf{82} & 80 \\
3 & put\_glasses & 67 & 50 & 62 & 50 & 67 & 50 \\
4 & put\_ring & 74 & 0 & 63 & 10 & \textbf{71} & 50 \\
5 & put\_items\_drw & 1 & 0 & \textbf{34} & 10 & 0 & 0 \\
6 & put\_blocks & 83 & 70 & \textbf{90} & 90 & 55 & 20 \\
7 & pour\_water & 19 & 0 & \textbf{22} & 0 & 12 & 0 \\
8 & insert\_wire & 18 & 0 & \textbf{32} & 0 & 25 & 0 \\
9 & pick\_items & 67.5 & 30 & \textbf{89.75} & 60 & 76.75 & 50 \\
10 & put\_stationery & 12 & 0 & 26 & 0 & \textbf{29} & 0 \\
\midrule
\multicolumn{8}{l}{\textit{Semantic Reasoning}} \\
\midrule
11 & sort\_headphone & 41 & 20 & 58 & 50 & \textbf{62} & 50 \\
12 & classify\_items & 55 & 10 & 70 & 30 & \textbf{94} & 80 \\
13 & press\_button & 31 & 0 & 34 & 0 & \textbf{34} & 0 \\
14 & pair\_up\_items & \textbf{77} & 60 & 70.5 & 30 & 80 & 50 \\
15 & pick\_fruits & 78 & 50 & \textbf{100} & 100 & \textbf{100} & 100 \\
\midrule
\multicolumn{2}{l}{\textbf{Total (/1500)}} & \textbf{702.5} & 20.0 & \textbf{867.25} & 32.7 & \textbf{801.75} & 35.3 \\
\bottomrule
\end{tabular}
\end{table*}

\section{Data scaling details}
\label{app:data_scaling_detail}
Table~\ref{tab:data_scaling_detail} reports the per-task score and success rate for each data recipe.

\begin{table*}[t]
\centering
\scriptsize
\caption{Detailed data-scaling results for the unified $\pi_{0.5}$ policy on five representative tasks. Entries are score\,/\,SR\,(\%); Total SR is averaged over the five evaluated tasks. Best score per task is shown in \textbf{bold}.}
\label{tab:data_scaling_detail}
\setlength{\tabcolsep}{3pt}
\renewcommand{\arraystretch}{1.15}
\begin{tabular}{@{}l cc cc cc cc cc cc@{}}
\toprule
& \multicolumn{2}{c}{\texttt{glasses}} & \multicolumn{2}{c}{\texttt{blocks}} & \multicolumn{2}{c}{\texttt{wire}} & \multicolumn{2}{c}{\texttt{button}} & \multicolumn{2}{c}{\texttt{fruits}} & \multicolumn{2}{c}{Total} \\
\cmidrule(lr){2-3} \cmidrule(lr){4-5} \cmidrule(lr){6-7} \cmidrule(lr){8-9} \cmidrule(lr){10-11} \cmidrule(lr){12-13}
Data & Sc & SR & Sc & SR & Sc & SR & Sc & SR & Sc & SR & Sc & SR \\
\midrule
50/task  & 29 & 0  & 16 & 0  & 21 & 0  & 22 & 0  & 77 & 50 & 165 & 10 \\
100/task & 47 & 0  & 66 & 50 & 29 & 0  & 19 & 0  & 66 & 30 & 227 & 16 \\
200/task & 69 & 50 & 76 & 60 & 24 & 0  & \textbf{54} & 20 & 89 & 80 & \textbf{312} & \textbf{42} \\
300/task & 35 & 20 & 73 & 70 & 19 & 0  & 37 & 10 & 81 & 50 & 245 & 30 \\
400/task & \textbf{74} & 70 & 43 & 20 & 26 & 0  & 40 & 0  & \textbf{93} & 90 & 276 & 36 \\
Full     & 67 & 50 & 83 & 70 & 18 & 0  & 31 & 0  & 78 & 50 & 277 & 34 \\
Full\textsuperscript{\dag} & 49 & 30 & \textbf{93} & 90 & \textbf{33} & 0 & 27 & 0 & 46 & 10 & 248 & 26 \\
\bottomrule
\end{tabular}
\\[2pt]
\footnotesize \textsuperscript{\dag}Full-data model trained for 120k steps; all other rows use 60k steps.
\end{table*}

\section{Real2Sim transfer details}
\label{app:real2sim_detail}

Table~\ref{tab:real2sim} reports the real-robot evaluation of checkpoints
trained with real demonstrations, simulated demonstrations, and a 50\% mixed
real/sim recipe. The paired simulation extension is used as a diagnostic data
source; real-robot trials remain the endpoint for capability assessment.

\begin{table}[t]
\centering
\small
\caption{Real2Sim transfer ($\pi_{0.5}$). Three checkpoints evaluated on the real robot. SR in \%.}
\label{tab:real2sim}
\setlength{\tabcolsep}{4pt}
\begin{tabular}{@{}l|cc|cc|cc@{}}
\toprule
& \multicolumn{2}{c|}{Real ckpt} & \multicolumn{2}{c|}{Sim ckpt} & \multicolumn{2}{c}{Cotrain 50\%} \\
Task & Sc & SR & Sc & SR & Sc & SR \\
\midrule
put\_blocks & 83 & 70 & \textbf{85} & 70 & \textbf{85} & 70 \\
press\_button & 31 & 0 & \textbf{65} & 50 & 40 & 20 \\
pick\_fruits & \textbf{78} & 50 & 51 & 30 & 68 & 50 \\
\bottomrule
\end{tabular}
\end{table}

\section{Per-task dataset statistics}
\label{app:dataset_stats}

Table~\ref{tab:dataset_stats} summarizes the collected demonstrations for each task, including trajectory count, average duration, and the primary manipulation capability being evaluated.

\begin{table*}[h]
\centering
\small
\caption{Per-task dataset statistics. Durations are averaged over collected trajectories.}
\label{tab:dataset_stats}
\setlength{\tabcolsep}{4pt}
\begin{tabular}{@{}cllrrrl@{}}
\toprule
\# & Task & Family & Traj. & Avg. frames & Avg. dur. & Primary capability \\
\midrule
1 & \texttt{arrange\_cup} & Execution & 528 & 1,056 & 52.8s & Bimanual stacking and placement \\
2 & \texttt{put\_spoon} & Execution & 525 & 543 & 27.1s & Object placement into container \\
3 & \texttt{put\_glasses} & Execution & 513 & 684 & 34.2s & Bimanual shelf placement \\
4 & \texttt{put\_ring} & Execution & 517 & 537 & 26.9s & Ring threading and alignment \\
5 & \texttt{put\_items\_drawer} & Execution & 510 & 1,143 & 57.2s & Drawer opening, placement, closure \\
6 & \texttt{put\_blocks} & Execution & 451 & 607 & 30.4s & Color-conditioned placement \\
7 & \texttt{pour\_water} & Execution & 526 & 1,035 & 51.8s & Pouring and container handling \\
8 & \texttt{insert\_wireline} & Execution & 530 & 697 & 34.9s & Cable handoff and insertion \\
9 & \texttt{pick\_items} & Execution & 532 & 485 & 24.3s & Repeated object transport \\
10 & \texttt{put\_stationery} & Execution & 525 & 962 & 48.1s & Zipper and container manipulation \\
11 & \texttt{sort\_headphone} & Semantic & 515 & 418 & 20.9s & Category-based selection \\
12 & \texttt{classify\_items} & Semantic & 545 & 701 & 35.1s & Shape classification \\
13 & \texttt{press\_button} & Semantic & 538 & 367 & 18.4s & Ordered button pressing \\
14 & \texttt{pair\_up\_items} & Semantic & 540 & 508 & 25.4s & Relational matching \\
15 & \texttt{pick\_fruits} & Semantic & 645 & 497 & 24.9s & Semantic object selection \\
16 & \texttt{put\_clothes} & Mobile & 540 & 3,871 & 193.6s & Navigation-conditioned placement \\
17 & \texttt{hang\_picture} & Mobile & 576 & 2,851 & 142.6s & Mobile bimanual hanging \\
18 & \texttt{organize\_shoes} & Mobile & 595 & 1,928 & 96.4s & Mobile pickup and arrangement \\
19 & \texttt{put\_bottle} & Mobile & 630 & 1,873 & 93.7s & Mobile shelf placement \\
20 & \texttt{set\_tableware} & Mobile & 531 & 4,220 & 211.0s & Long-horizon table setting \\
\bottomrule
\end{tabular}
\end{table*}

\section{Data format and observation/action layout}
\label{app:data_format}

ManipArena trajectories are stored in LeRobot v2.1 format. Tabletop tasks use 56-dimensional state and action vectors. Mobile manipulation tasks add 6 dimensions for head, lifting column, and chassis, giving 62D in total. The end-effector frame uses $+x$ forward, $+y$ left, $+z$ up. The last element of each 7D joint group (indices 20, 27, 34, 41, 48, 55) is the gripper channel. Table~\ref{tab:data_layout} lists the full per-dimension layout.

\begin{table}[h]
\centering
\small
\caption{Per-dimension layout of ManipArena state and action vectors. Tabletop tasks use indices 0--55 (56D); mobile tasks additionally use indices 56--61 (62D).}
\label{tab:data_layout}
\setlength{\tabcolsep}{4pt}
\begin{tabular}{@{}c l c l@{}}
\toprule
Index & Field & Dim & Description \\
\midrule
\multicolumn{4}{l}{\textit{End-effector (14D)}} \\
\midrule
0--2  & \texttt{follow\_left\_ee\_cartesian\_pos}  & 3 & Left arm position $(x, y, z)$ \\
3--5  & \texttt{follow\_left\_ee\_rotation}        & 3 & Left arm rotation (roll, pitch, yaw) \\
6     & \texttt{follow\_left\_gripper}             & 1 & Left gripper open/close \\
7--9  & \texttt{follow\_right\_ee\_cartesian\_pos} & 3 & Right arm position $(x, y, z)$ \\
10--12 & \texttt{follow\_right\_ee\_rotation}      & 3 & Right arm rotation (roll, pitch, yaw) \\
13    & \texttt{follow\_right\_gripper}            & 1 & Right gripper open/close \\
\midrule
\multicolumn{4}{l}{\textit{Left arm joints (21D)}} \\
\midrule
14--20 & \texttt{follow\_left\_arm\_joint\_pos}  & 7 & Joint positions (6 joints + gripper) \\
21--27 & \texttt{follow\_left\_arm\_joint\_dev}  & 7 & Joint velocities (6 joints + gripper) \\
28--34 & \texttt{follow\_left\_arm\_joint\_cur}  & 7 & Joint currents (6 joints + gripper) \\
\midrule
\multicolumn{4}{l}{\textit{Right arm joints (21D)}} \\
\midrule
35--41 & \texttt{follow\_right\_arm\_joint\_pos} & 7 & Joint positions (6 joints + gripper) \\
42--48 & \texttt{follow\_right\_arm\_joint\_dev} & 7 & Joint velocities (6 joints + gripper) \\
49--55 & \texttt{follow\_right\_arm\_joint\_cur} & 7 & Joint currents (6 joints + gripper) \\
\midrule
\multicolumn{4}{l}{\textit{Mobile extras (mobile tasks only, 6D)}} \\
\midrule
56--57 & \texttt{head\_actions}                & 2 & Head rotation (yaw, pitch) \\
58     & \texttt{height}                       & 1 & Lifting-column height \\
59--61 & \texttt{velocity\_decomposed\_odom}    & 3 & Chassis velocity $(v_x, v_y, \omega)$ \\
\bottomrule
\end{tabular}
\end{table}

ManipArena additionally releases joint velocities and joint currents per timestep (indices 21--34, 42--55), providing additional signals related to contact and actuation dynamics that are useful for compliant- or impedance-based control policies. For mobile trials, the head, lifting-column, and chassis channels are released alongside the arm channels so that navigation-conditioned manipulation policies can be trained and evaluated on a uniform observation/action interface.

\section{Evaluation protocol details}
\label{app:protocol}

This appendix collects the formal procedures referenced from Section~\ref{sec:benchmark}.

\paragraph{Schema specification and sampling.}
The training-time distribution in Equation~\ref{eq:schema} is given by the per-task schema $\mathcal{S}_i = \{(D_{i,j}, p_{i,j})\}_{j=1}^{M_i}$, where $D_{i,j}$ is the $j$-th diversity dimension and $p_{i,j}$ is a categorical distribution over its admissible values. In the common case, $P_i^{\mathrm{train}}(c) = \prod_{j=1}^{M_i} p_{i,j}(d_{i,j})\,p_i^{\delta}(\delta)$. Other tasks use joint distributions over subsets of dimensions (e.g., enumerated 4-item combinations in \texttt{pick\_items\_basket}, fixed orderings in \texttt{press\_button}) or refresh certain dimensions only at batch boundaries (the \texttt{change\_frequency} field). The full per-task schema (diversity dimensions, distributions, and constraints) is released alongside the dataset. Algorithm~\ref{alg:collect} summarizes the data-collection loop.

\begin{algorithm}[h]
\caption{Schema-driven data collection for task $i$}
\label{alg:collect}
\begin{algorithmic}[1]
\Require Schema $\mathcal{S}_i = \{(D_{i,j}, p_{i,j})\}_{j=1}^{M_i}$; target trajectory count $N_i$
\Ensure Trajectory set $\mathcal{D}_i$
\State $\mathcal{D}_i \gets \emptyset$
\While{$|\mathcal{D}_i| < N_i$}
    \State Sample a configuration $c = (d_{i,1}, \dots, d_{i,M_i}, \delta) \sim \mathcal{S}_i$, where $\delta$ is the distractor condition specified by the schema
    \State Reset scene to $c$
    \State Collect teleoperated demonstration to obtain trajectory $\tau$
    \State Generate three-level annotation $\ell = (\ell_{L1}, \ell_{L2}, \ell_{L3})$
    \State $\mathcal{D}_i \gets \mathcal{D}_i \cup \{(\tau, \ell)\}$
\EndWhile
\State \Return $\mathcal{D}_i$
\end{algorithmic}
\end{algorithm}

\paragraph{Per-band aggregation.}
The stratified distributions in Equation~\ref{eq:strat}, the per-trial score in Equation~\ref{eq:score}, and the aggregate metrics in Equation~\ref{eq:agg_sr} are used for all real-robot results in Section~\ref{sec:experiments}. Per-band score means provide a finer-grained breakdown than the cross-task aggregate:
\begin{equation}
\label{eq:band}
\bar S_i^{B} \;=\; \frac{1}{|B|}\sum_{t \in B} S_i^{(t)}, \qquad B \in \{T_{1\text{-}4},\, T_{5\text{-}8},\, T_{9\text{-}10}\}.
\end{equation}
The single-task success rate is the special case $\mathrm{SR}_i = \mathrm{SR}(\{i\})$ of Equation~\ref{eq:agg_sr} and counts a trial with score $\geq 9/10$ as a success, providing a near-completion signal alongside the partial-credit total.

\begin{algorithm}[h]
\caption{Stratified 10-trial evaluation of policy $\pi$ on task $i$}
\label{alg:eval}
\begin{algorithmic}[1]
\Require Policy $\pi$; schema $\mathcal{S}_i$; rubric $\{(w_{i,k}, g_{i,k})\}_{k=1}^{K_i}$
\Ensure Per-trial scores $\{S_i^{(t)}\}_{t=1}^{10}$
\For{$t = 1, \dots, 10$}
    \If{$t \leq 4$} \Comment{in-domain}
        \State Sample $c_t$ from the training-time support of $\mathcal{S}_i$
    \ElsIf{$t \leq 8$} \Comment{visual / configuration shift}
        \State Sample $c_t$ as a controlled visual or configuration shift specified by $\mathcal{S}_i$
    \Else \Comment{strongest held-out}
        \State Sample $c_t$ from the strongest available held-out condition (Appendix~\ref{app:ood_audit})
    \EndIf
    \State Reset scene to $c_t$ and execute $\pi$ to produce a trajectory
    \State $S_i^{(t)} \gets \sum_{k=1}^{K_i} w_{i,k}\,\mathbf{1}[g_{i,k}\text{ achieved}]$
\EndFor
\State \Return $\{S_i^{(t)}\}_{t=1}^{10}$
\end{algorithmic}
\end{algorithm}

\section{Scoring rubric summary}
\label{app:rubrics}

\begin{table*}[h]
\centering
\small
\caption{Task scoring rubrics used for partial-credit evaluation. Each trial is worth 10 points.}
\label{tab:rubrics}
\setlength{\tabcolsep}{4pt}
\begin{tabular}{@{}p{0.22\textwidth}p{0.72\textwidth}@{}}
\toprule
Task & Scoring stages \\
\midrule
\texttt{arrange\_cup} & Pick cup1 (1), place cup1 (1), pick cup2 (1), place cup2 (2), pick cup3 (1), stack apex cup (3), retract (1) \\
\texttt{put\_spoon} & Pick bowl (2), place bowl center (2), pick spoon (2), place spoon in bowl (3), retract (1) \\
\texttt{put\_glasses} & Pick glasses (2), place center (2), adjust orientation (2), bimanual shelf placement (3), retract (1) \\
\texttt{put\_ring} & Move rod to center (3), pick ring (3), thread ring onto rod (3), retract (1) \\
\texttt{put\_items\_drawer} & Three repetitions of open drawer (1), place item (1), close drawer (1), plus retract (1) \\
\texttt{put\_blocks} & Place red block (3), place yellow block (3), place green block (3), retract (1) \\
\texttt{pour\_water} & Move cup to center (2), grasp kettle (2), pour (3), place kettle back (2), retract (1) \\
\texttt{insert\_wireline} & Left hand picks cable (3), right hand receives cable (3), insert into socket (3), retract (1) \\
\texttt{pick\_items} & Four repetitions of pick item (1) and place in basket (1.25), plus retract (1) \\
\texttt{put\_stationery} & Hold case (1), unzip (1.5), open (1), place three items (3), hold case (1), zip (1.5), retract (1) \\
\texttt{sort\_headphone} & Find headphone in clutter (2), pick up (3), place in basket (4), retract (1) \\
\texttt{classify\_items} & Arrange first item (3), arrange second item (3), arrange third item (3), retract (1) \\
\texttt{press\_button} & Press first button (3), press second button (3), press third button (3), retract (1) \\
\texttt{pair\_up\_items} & Pick match for upper-left item (2), place below (2.5), pick match for upper-right item (2), place below (2.5), retract (1) \\
\texttt{pick\_fruits} & Pick each target fruit with weights normalized by target count; wrong picks receive a one-point penalty; retract completes the trial \\
\texttt{put\_clothes} & Navigation and manipulation stages for shirt, pants, and socks: navigate, pick, transport, place, and final retraction \\
\texttt{hang\_picture} & Navigate, bimanual pick (3), navigate, raise picture (2), hang picture (2), retract \\
\texttt{organize\_shoes} & Navigate, pick left shoe (2), pick right shoe (2), navigate, place left shoe (1.5), place right shoe (1.5), retract \\
\texttt{put\_bottle} & Navigate (2), pick bottle (2), navigate (2), place on shelf (2), retract (2) \\
\texttt{set\_tableware} & Navigate, pick basket (1), navigate, place basket, place plate, place knife (2), place fork (2), retract \\
\bottomrule
\end{tabular}
\end{table*}

\section{Detailed limitations}
\label{app:limitations}

\paragraph{Real-robot stochasticity.}
ManipArena evaluates policies on physical robots, where repeated executions of
the same checkpoint on nominally identical trials can still vary due to contact
conditions, object pose perturbations, sensing noise, latency, controller
variation, and reset differences. The stratified 10-trial protocol and
partial-credit scoring reduce the effect of a single binary outcome, but the
reported scores should still be interpreted as empirical estimates under a
fixed protocol rather than deterministic checkpoint properties.

\paragraph{Cost and coverage.}
Real-robot evaluation requires robot time, scene reset, failure recovery, and
hardware supervision; the controlled studies also require training multiple
checkpoints across data recipes, language fields, and model families. This cost
limits the breadth of the current sweep. The tabletop model zoo includes seven
configurations, mobile policy evaluation is limited to one task-specific
$\pi_{0.5}$ configuration on the five mobile tasks, and the paired simulation
extension covers three tabletop tasks. These choices provide
controlled evidence for the factors studied in the main paper, but do not
exhaust all robot foundation models, training schedules, prompt formats, data
mixtures, or real/sim recipes.

\paragraph{Platform and environment scope.}
The controlled green-screen environments are designed to isolate evaluation
variables, but they do not cover the full visual and physical complexity of
open homes, factories, or outdoor environments. The current results are also
tied to the tested robot embodiments, object sets, reset procedures, and
scoring rubrics. Future instantiations can reuse the same task schemas and
partial-credit framework while varying scene, lighting, camera viewpoint,
object set, or robot embodiment.

\paragraph{Instruction interface.}
The cross-model evaluation uses a common high-level L1 prompt to keep the
test-time instruction interface fixed across VLA and WAM policies. This
supports controlled comparison, but it does not characterize every prompting
regime. Some policy families may benefit from more explicit scene-grounded or
action-level instructions. Systematically varying test-time prompt granularity
and model-specific instruction interfaces is therefore left for future work.

\section{OOD authenticity audit}
\label{app:ood_audit}

The 10-trial stratification (Section~\ref{sec:stratified_eval}) reserves trials T9--T10 for the strongest held-out condition available per task. Whether that condition introduces an unseen object depends on whether object identity is part of the task semantics. Table~\ref{tab:ood_audit} records the per-task assignment used in this paper.

\begin{table}[h]
\centering
\small
\caption{Per-task held-out condition for the 15 tabletop tasks. \emph{Object OOD} tasks introduce previously unseen objects at T9--T10. \emph{Configuration OOD} tasks hold out spatial, ordering, or procedural combinations from the training schema while reusing in-distribution objects.}
\label{tab:ood_audit}
\setlength{\tabcolsep}{6pt}
\begin{tabular}{@{}lll@{}}
\toprule
Task & T5--T8 shift & T9--T10 condition \\
\midrule
\multicolumn{3}{l}{\textit{Object OOD (7 tasks)}} \\
\midrule
\texttt{arrange\_cup} & mug $\to$ paper cup & plastic cup (blue) \\
\texttt{sort\_headphone} & wired $\to$ over-ear $\to$ Bluetooth & neckband sport (white) \\
\texttt{put\_spoon} & stainless $\to$ children's spoon & black plastic spoon \\
\texttt{put\_glasses} & adult $\to$ children's $\to$ goggles & sunglasses \\
\texttt{put\_items\_drawer} & object-combination change & marker, watch, top \\
\texttt{put\_stationery} & color-combination change & fountain pens \\
\texttt{pick\_fruits} & banana introduced + new basket & orange/cherry + toy doll \\
\midrule
\multicolumn{3}{l}{\textit{Configuration OOD (8 tasks)}} \\
\midrule
\texttt{insert\_wireline} & white $\to$ black cable & rare position combinations \\
\texttt{put\_blocks} & block-size combinations & rare color orderings \\
\texttt{pour\_water} & cup-type combinations & rare cup-position pairings \\
\texttt{put\_ring} & ring-position variation & rare ring/rod-position pairings \\
\texttt{press\_button} & button-order permutations & held-out button orders \\
\texttt{classify\_items} & shape-order permutations & held-out shape orders \\
\texttt{pair\_up\_items} & matching-pair combinations & held-out match pairs \\
\texttt{pick\_items\_basket} & basket position + item subset & basket-left configurations \\
\bottomrule
\end{tabular}
\end{table}

For the 8 Configuration OOD tasks, T9--T10 stresses spatial, ordering, or procedural generalization rather than object semantics. As discussed in Section~\ref{sec:stratified_eval}, this separation is intentional. Future versions of the benchmark can introduce object OOD for the remaining 8 tasks (e.g., triangular rings for \texttt{put\_ring}, orange or purple buttons for \texttt{press\_button}) without altering the partial-credit framework or the controlled environment.

\end{document}